\documentclass[%
 aip,
 amsmath,amssymb,
 reprint,%
]{revtex4-1}
\usepackage[english]{babel}
\usepackage{graphicx}
\usepackage{dcolumn}
\usepackage{bm}

\usepackage[utf8]{inputenc}
\usepackage[T1]{fontenc}
\usepackage{mathptmx}
\usepackage{etoolbox}
\usepackage{xcolor}

\makeatletter
\def\@email#1#2{%
 \endgroup
 \patchcmd{\titleblock@produce}
  {\frontmatter@RRAPformat}
  {\frontmatter@RRAPformat{\produce@RRAP{*#1\href{mailto:#2}{#2}}}\frontmatter@RRAPformat}
  {}{}
}%
\makeatother
\begin{document}

\preprint{AIP/123-QED}

\title[Denoising and Reconstruction of Nonlinear Dynamics using Truncated Reservoir Computing]{Denoising and Reconstruction of Nonlinear Dynamics using Truncated Reservoir Computing}
\author{Omid Sedehi}
\email{omid.sedehi@uts.edu.au}
\homepage{Corresponding Author}
\affiliation{Centre for Audio, Acoustics and Vibration (CAAV), School of Mechanical and Mechatronic Engineering, University of Technology Sydney, Ultimo, NSW 2007, Australia}

\author{Manish Yadav}%
\email{manish.yadav@tu-berlin.de}
\affiliation{Chair of Cyber-Physical Systems in Mechanical Engineering, Technische Universität Berlin, Straße des 17. Juni 135, 10623 Berlin, Germany}%
 
\author{Merten Stender}
\email{merten.stender@tu-berlin.de}
\affiliation{Chair of Cyber-Physical Systems in Mechanical Engineering, Technische Universität Berlin, Straße des 17. Juni 135, 10623 Berlin, Germany}%

\author{Sebastian Oberst}
\email{sebastian.oberst@uts.edu.au}
\affiliation{Centre for Audio, Acoustics and Vibration (CAAV), School of Mechanical and Mechatronic Engineering, University of Technology Sydney, Ultimo, NSW 2007, Australia}

\date{\today}

\begin{abstract}
\textbf{Abstract}
\\
Measurements acquired from distributed physical systems are often sparse and noisy. Therefore, signal processing and system identification tools are required to mitigate noise effects and reconstruct unobserved dynamics from limited sensor data. However, this process is particularly challenging because the fundamental equations governing the dynamics are largely unavailable in practice. Reservoir Computing (RC) techniques have shown promise in efficiently simulating dynamical systems through an unstructured and efficient computation graph comprising a set of neurons with random connectivity. However, the potential of RC to operate in noisy regimes and distinguish noise from the primary smooth or non-smooth deterministic dynamics of the system has not been fully explored. This paper presents a novel RC method for noise filtering and reconstructing unobserved nonlinear dynamics, offering a novel learning protocol associated with hyperparameter optimization. The performance of the RC in terms of noise intensity, noise frequency content, and drastic shifts in dynamical parameters is studied in two illustrative examples involving the nonlinear dynamics of the Lorenz attractor and the adaptive exponential integrate-and-fire system. It is demonstrated that denoising performance improves by truncating redundant nodes and edges of the reservoir, as well as by properly optimizing hyperparameters, such as the leakage rate, spectral radius, input connectivity, and ridge regression parameter. Furthermore, the presented framework shows good generalization behavior when tested for reconstructing unseen and qualitatively different attractors. Compared to the extended Kalman filter, the presented RC framework yields competitive accuracy at low signal-to-noise ratios and high-frequency ranges.
\end{abstract}

\maketitle

\begin{quotation}
Reconstruction of unobserved dynamical responses from sparse and noisy observations is a crucial and challenging task when dealing with unknown nonlinear systems. Filtering techniques have commonly been employed to handle such problems, but they often demand a thorough understanding of the underlying processes, such as the governing equations in terms of a physics-based state-space model. This paper presents a novel truncated reservoir computing approach to simulate nonlinear dynamics and implicitly recover the underlying physics from noisy measurements. It is proposed that reservoir computers can effectively distinguish noise from dynamics and reconstruct unmeasured dynamical responses from noisy ones without having to involve any supplementary noise reduction techniques. While the RC acts as a model-free mathematical modeling technique, it offers competitive accuracy compared to the conventional filtering techniques at certain noise levels.
\end{quotation}

\section{\label{sec:introduction}Introduction}
Measurements of physical quantities from distributed systems are often noisy and incomplete in covering all response quantities. This issue has inspired the development of a wide range of denoising and reconstruction methods, which fall under two major classes, namely model-free and model-based methods, depending on whether they demand prior knowledge of underlying physics. A good example of the former is wavelet-based denoising \cite{Donho1995}, which does not require incorporating physical knowledge explicitly, whereas stochastic filtering techniques, including Kalman-type filters \cite{Anderson1979}, often demand a general understanding of the physics, characterized via state-space models. However, the foundation of both methods primarily rests upon a key hypothesis, implying that the desired signal and the noise have distinguishable low and high-dimensional features in a specific latent space.

Reservoir computing is a computational framework rooted in the broader domain of recurrent neural networks (RNNs) and dynamical systems theory. It leverages the dynamics of a fixed, randomly initialized high-dimensional system, known as the ``reservoir,'' to transform sequential input data into a latent space representation. In this paradigm, only the output layer is trained through ridge regression \cite{Ghani2010}, which linearly combines the reservoir states to minimize prediction errors. This significantly reduces the computational complexity associated with the training, mainly because only a system of linear equations needs to be solved coupled with a proper regularization. RC has been particularly effective in time-series processing \cite{gauthier2021, Throne2022, Shahi2022}, speech recognition \cite{Ghani2010, Araujo2020}, and chaotic system modeling \cite{Pathak2017, Krishnagopal2020, Carroll2022} due to its ability to exploit the reservoir's rich temporal dynamics and memory capacity. Heuristics of building the reservoir formulation, e.g., including an output feedback or a delayed input, have been shown to improve the results \cite{Jaurigue2021}. The generalization behavior of RC has recently been studied, where an RC approach is employed for making predictions on unexplored attractors \cite{Flynn2021, Parlitz2024}. However, the optimal size and structure of the reservoir remain an open question since reservoir networks are mostly initialized considering random connectivity. Only lately researchers have started investigating the structure-function relationships to discover network properties that are beneficial for a given learning task within the RC framework \cite{Manish2024, Yadav2024}.

Recent research has explored the impact of noise on the RC's performance, revealing its strength and weaknesses for signal reconstruction across various contexts. Rarely RC's have been employed specifically for simultaneous denoising and signal reconstruction tasks. Semenova et al. \cite{Semenova2019} have studied the propagation of noise from the input nodes toward the reservoir outputs in recurrent and multilayer networks. Estébanez et al. \cite{Estébanez2019} have shown that, while noise can degrade the performance of RC-based analog computations, its introduction in the training phase can lead to improved robustness and accuracy. Similarly, Wikner et al. \cite{WIKNER2024} have shown that adding noise to the training data can improve the stability of predictions. Röhm et al. \cite{Rohm2021} have shown that an RC trained based on a single noisy trajectory may still be able to reproduce unexplored attractors. Nathe et al. \cite{Nathe2023} have investigated the impact of measurement noise on the performance of RC when used for reconstructing chaotic systems. Lee et al. \cite{Lee2024} have employed RC along with a nonlinear forward operator for efficient image denoising. Choi and Kim \cite{Choi2024} have separated the signal from noise as the predictable part modeled by RC and acquired information about the additive or multiplicative nature of the noise, as well as noise statistical properties. Despite these advances in RC-based denoising approaches, a thorough investigation is still required to explore the potential of RC for discarding different noise processes and factors affecting denoising performance.

This work introduces a novel learning protocol for denoising and reconstruction of nonlinear dynamics based on a truncated RC framework. This approach begins with training the reservoir on a specific data set using the conventional ridge regression, followed by a new optimization strategy that calibrates key hyperparameters, including the ridge regression coefficient, leakage rate, and spectral radius, aiming to enhance performance. Once the optimized reservoir is obtained, we truncate redundant nodes and edges according to the model accuracy on the validation dataset using a novel pruning strategy. Two numerical examples are provided to demonstrate the proposed protocol and study the impact of noise intensity and frequency content on the performance of the RC. The generalization behavior of the proposed networks is demonstrated against drastic changes in the dynamical parameters and Signal-to-Noise Ratio (SNR). Comparison with the Extended Kalman Filter (EKF), a near optimal filter \cite{Anderson1979}, is provided for the Lorenz attractor example, considering different noise levels. Finally, the method is further demonstrated using the non-smooth deterministic dynamics of the adaptive exponential integrate-and-fire (AdEx) model.

This study primarily investigates the potential of reservoir computing for denoising and state reconstruction. In addition, it applies hyperparameter optimization and explores the pruning of redundant nodes and edges. The pruning of small reservoir networks, specifically the removal of redundant nodes and connections, has received relatively little attention in previous studies. Notably, Dutoit et al \cite{Dutoit2009} have shown that the pruning of the readout nodes is an effective strategy to improve the prediction accuracy and reduce hardware and software computational demands. Scardapane et al. \cite{Scardapane2015} have characterized the significance of reservoir synapses according to the correlation between the input and output of the neurons and specified the significance of neurons based on the importance of their neighboring synapses. Then, based on this criterion, they have presented a pruning approach for reducing the reservoir dimensionality without calculating the error gradients \cite{Scardapane2015}. Haluszczynski et al. \cite{Haluszczynski2020} have shown that removing the nodes with negative impact on the reservoir's response to the input can enhance the accuracy and reduce instabilities. Compared to these works, this study provides a more systematic and generic approach to reducing the reservoir nodes and edges.

The remainder of this paper is organized as follows: first, the mathematical foundation of the RC framework will be reviewed, presenting the conventional ridge regression method. Then, the reservoir's hyperparameters will be identified, for which an optimization approach will be presented. Truncation of the RC through a pruning approach will be discussed, and further comparisons will be made between the proposed RC framework and conventional filtering techniques from a theoretical standpoint. The presented approaches will be demonstrated using two nonlinear systems, aiming to address simultaneous state reconstruction and denoising tasks. Finally, further discussions and conclusions will be drawn about the denoising performance of the proposed truncated RC framework.

\section{\label{sec:methods}Reservoir Computing Framework}
\subsection{Echo State Network} 
Several neural network architectures are available for constructing reservoir computers, among which Echo State Network (ESN) \cite{Jaeger2007} and Liquid State Machines \cite{Maass2002, Hazan2012} are perhaps the most widely used ones. In this paper, ESN is employed for creating a reservoir whose formulation is described in terms of the latent dynamical states:
\begin{equation}
\mathbf{r}(t_{i+1}) = (1-{\it{\alpha}})\mathbf{r}(t_i)+{\it{\alpha}} \ f(\mathbf{W}_{\mathrm{res}} \mathbf{r}(t_i)+\mathbf{W}_{\mathrm{in}} \mathbf{u}(t_i) + \mathbf{b})
\label{EQ1}
\end{equation}
where $f(.)$ is the activation function, considered $\tanh(.)$, $\mathbf{r}(t_i)$ is the reservoir state at time $t_i$, $\alpha$ is the leakage rate representing how much of the preceding state contributes to the present state, $\mathbf{W}_{\mathrm{res}}$ is the reservoir weight matrix, considered a random weighted Erdős–Rényi (ER) graph matrix with $p$ internal connectivity probability, $\mathbf{W}_{\mathrm{in}}$ is the input weight matrix scaled by a factor of $\zeta$, $\mathbf{u}(t_i)$ is the input vector at time $t_i$, and \(\mathbf{b}\) is the bias term. Notably, an important parameter of the reservoir matrix ($\mathbf{W}_{\mathrm{res}}$) is its spectral radius, denoted by $\gamma$ and calculated as the greatest eigenvalue of the reservoir matrix \cite{Ren2022}.

Given the above equation, a recursive formulation is available for calculating the reservoir states. By collecting all states into a matrix like $\mathbf{R}_n = [\mathbf{r}(t_1),...,\mathbf{r}(t_n)]^T$, simulated responses of the physical system at different time steps comprised in $\bold{Y}_n = [\mathbf{y}(t_1),...,\mathbf{y}(t_n)]^T$ can be described as
\begin{equation}
\mathbf{Y_{\it{n}}} = \mathbf{R_{\it{n}}}\mathbf{W}_{\mathrm{out}} 
\label{EQ2}
\end{equation}
In the training phase of the RC framework, it is customarily assumed that the reservoir is fixed, whereas the readout matrix $\mathbf{W}_{\mathrm{out}}$ must be trained. Thus, the optimization will simplify into a linear regression, which requires minimizing the following loss function:
\begin{equation}
L = \|\mathbf{{\hat{Y}_{\it{n}}} - \mathbf{R_{\it{n}}}\mathbf{W}_{\mathrm{out}}}\|_2^2 + \lambda \|\mathbf{W}_{\mathrm{out}}\|_2^2
\label{EQ3}
\end{equation}
where $\mathbf{\hat{Y}_{\it{n}}}$ is a vector of actual measurements fed into the RC as input. This formulation, referred to as linear ridge regression \cite{Hoerl1970}, can readily calibrate the readout nodes through an explicit solution, i.e.,
\begin{equation}
\mathbf{\hat{W}}_{\mathrm{out}} = (\mathbf{R^{\it{T}}_{\it{n}}}\mathbf{R_{\it{n}}}+\it{\lambda} \bold{I})^{-1}\mathbf{R_{\it{n}}^{\it{T}}} \mathbf{\hat{Y}_{\it{n}}} 
\label{EQ4}
\end{equation}
where ${\hat{\mathbf{W}}}_{\mathrm{out}}$ is the optimal readout matrix, and $\lambda$ is the ridge coefficient, often selected to be a small coefficient like $10^{-8}$. In Eq. \ref{EQ3}, it is evident that the first term, $\|{{{\mathbf{\hat{Y}_{\it{n}}}}} - \mathbf{R_{\it{n}}}\mathbf{W}_{\mathrm{out}}}\|_2^2$, provides a measure of fitting accuracy, whereas the second term, $ \lambda \|\mathbf{W}_{\mathrm{out}}\|_2^2$, involves a penalty over the non-zero elements of $\mathbf{W}_{\mathrm{out}}$, implementing a smooth sparsity. Note that in problems involving different levels of noise, $\lambda$ must be inferred properly based on the data.

\subsection{Proposed Learning Protocol}
\begin{figure}
    \centering
    \includegraphics[width=1.0\linewidth]{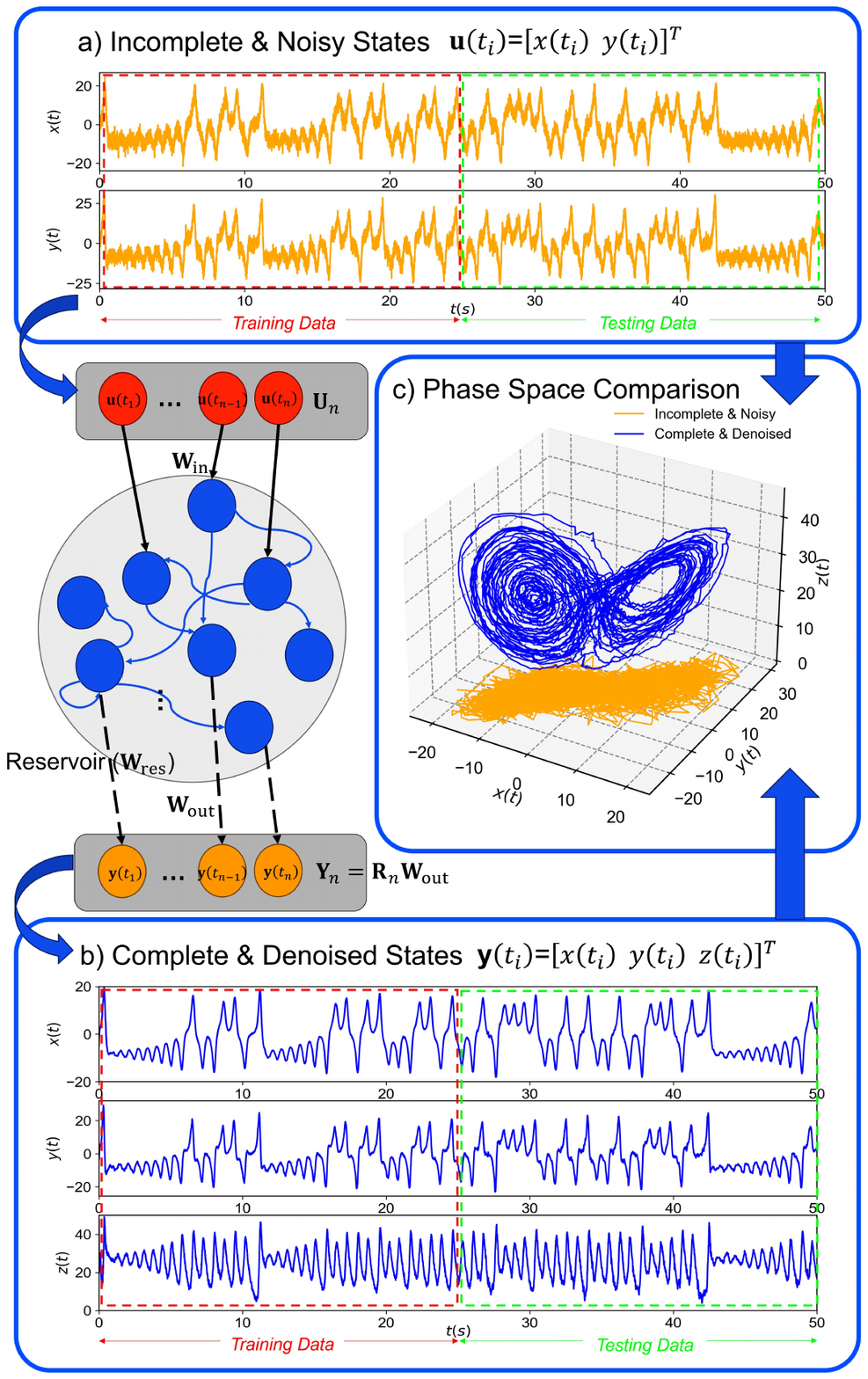}
    \caption{Overview of the proposed RC protocol for denoising and reconstruction of nonlinear dynamics. Blue, red, and orange circles show reservoir nodes, input nodes, and readout nodes, respectively. The blue arrows show the edges of the reservoir chosen randomly. The solid black arrows show the input connection to specific nodes of the reservoir, and the black dashed arrows show the readout connections. (a) Incomplete and noisy dynamics of the Lorenz system considered as the input to the RC (b) Reproduced Lorenz dynamics obtained through the RC framework (c) Comparison of the input and output of the reservoir in the phase space, representing how three-dimensional dynamics (blue curve) are reconstructed from two-dimensional noisy signals (orange curve).}
    \label{fig:FIG. 1}
\end{figure}

The reservoir is assumed to take noisy and incomplete measured data $\mathbf{U_{\it{n}}} =  [\mathbf{u}(t_1),...,\mathbf{u}(t_n)]^T$ as the input and return filtered and complete states ${\mathbf{Y_{\it{n}}}}$ as the output. FIG. \ref{fig:FIG. 1} provides an overview of this specific implementation of the reservoir. At first glance, it might be difficult to train such an RC since both noisy and clean data are required in the training phase. However, such data can be obtained via computer simulations wherein the dynamics are considered entirely known for specific chaotic orbits. Another possibility is to employ simpler system dynamics for acquiring noisy data and attempting to clean those measurements through conventional filtering techniques \cite{Donho1995}. In both cases, the RC trained based on these data sets is anticipated to generalize across a large neighborhood such that it can be employed for the denoising and reconstruction of unexplored nonlinear dynamics while being trained on limited clean data.

\subsection{Hyperparameter Optimization}
The RC framework presented in the preceding section is straightforward to follow and implement because it only requires performing a linear regression for the training of the readout unit. However, the performance of the reservoir computer often benefits from the fine-tuning of the reservoir and optimizing its underlying parameters. Such parameters include the number of neurons, leakage rate, spectral radius, input scaling, and node connectivity, whose optimization can be performed through a greedy algorithm that minimizes the loss function of Eq. \ref{EQ3} for the underlying parameters while considering a held-out data set. To formalize such an optimization, let the hyperparameters of the RC be denoted by $\phi$, specifically comprising the number of reservoir nodes ($N$), the leakage rate ($\alpha$), the spectral radius ($\gamma$), the input scaling ratio ($\zeta$), and the reservoir internal connectivity rate ($p$). The optimal values of these hyperparameters should be determined from
\begin{equation}
\mathbf{\hat{\phi}} = \arg\min_{\phi} \left(L\right) 
\label{EQ5}
\end{equation}
where the loss function is given by Eq. \ref{EQ3}. It should be noted that this minimization might encounter computational difficulties if gradient-descent-based optimization approaches \cite{Bottou1998} are employed. In effect, calculating the gradients requires implementing back-propagation \cite{Rumelhart1986} through the hidden states of the reservoir, which is often deemed challenging if not impossible, especially when the number of time steps in the training data is large \cite{Hanin2018}. Therefore, grid-based \cite{Shahi2022ML} or random search algorithms \cite{Kaelo2006} are often preferred in the literature. In this study, surrogate-assisted optimization techniques are employed along with cross-validation algorithms, which avoid computing the gradients \cite{Li2023} using the Python-based package {\it{scikit-optimize}}\cite{Bergstra2012}.

\subsection{Tuning the Ridge Coefficient}
The ridge regression coefficient ($\lambda$) becomes important when dealing with noisy data. This coefficient is inversely proportional to the level of noise contained in the data, and in the machine learning literature, it is often calibrated through cross-validation techniques \cite{Hastie2017}. It should be noted that this coefficient can also be treated as a hyperparameter and calibrated via Eq. \ref{EQ5}. However, due to the unknown nature of noise level and the wide range of potential values for this parameter, the inclusion of this parameter in the optimization algorithm might not be feasible. Therefore, an empirical optimization algorithm is preferred, wherein the optimal value of this parameter is selected from a relatively coarse one-dimensional grid. For this purpose, a set of exponents of ten is introduced to the cross-validation algorithm, i.e., $\{10^{-15},10^{-14},...,10^{20}\}$, and then, the best choice is specified through computing the cross-validation score. Although this approach might not be precise in terms of simultaneously optimizing all hyperparameters, it offers significant computational merits as it retains the closed-form solution of Eq. \ref{EQ3}.
 
\subsection{Pruning and Evolution Mechanisms}
The above optimization approach can be used to specify an optimal number of nodes for the RC while edges, representing the internal connectivity in the reservoir, are selected at random. Due to this arbitrary initialization, such an RC might not be in its minimal form, and there could be redundant nodes and edges, degrading the prediction performance. In this sense, it is reasonable to explore ways to reduce redundancies through truncating unnecessary nodes and edges \cite{Yadav2024, Manish2024}. Attaining minimality in the size and structure of the reservoir is also important for hardware development when a digital circuit is going to be designed for implementation via a physical RC \cite{Kent2024}. Therefore, to remove ineffective nodes and edges, one can first train the reservoir and its hyperparameters, considering a specific number of nodes and random edges, and then perform an exhaustive search for discarding a predefined percentage of nodes and connections, subject to satisfying a performance metric. The second stage can be implemented considering another optimization scheme, where the mean-squared error (MSE) calculated for the validation data set should be minimized while the pruning of nodes and edges takes place. However, this pruning strategy requires establishing a ranking mechanism across the nodes and edges so that the elimination can start from the least influential nodes. Five measures are considered for weighing the nodes' importance:
\begin{itemize}
    \item{Absolute mean state: The absolute average of the latent states is a metric to assess a node's activity over the entire training period. Higher average values indicate greater node activity.}
    \item {State variance: A higher variance in nodal states may indicate higher node activity.}
    \item {Number of edges: Nodes with more connections may have a greater influence on the RC's performance.}
    \item {Clustering: Nodes with a larger clustering coefficient can be more important as they better connect neighboring nodes.}
    \item {Page-rank: This web-inspired scoring approach integrates the influence of inbound and outbound edges, as well as the quality of edges as measured in terms of their relative weight, frequency of interaction, and contextual relevance within the network.\cite{Hagberg2008}.}
\end{itemize}
The above measures can be normalized to show a score in each category between zero and one. Then, the summation of these factors can be used for selecting the nodes nominated for elimination. Based on a preselected truncation percentage, redundant nodes will be eliminated, and the RC hyperparameters will be recalibrated. This process continues until a target performance is either achieved or the maximum number of trials is attained. Here, we hypothesize the usefulness of such metrics based on our intuition, but choosing other metrics for pruning nodes and edges is also possible. However, the decision whether to remove or retain the nominated nodes or edges must be made based on the performance.

In contrast to the pruning approach, we might encounter cases in which the number of nodes appears insufficient to describe the dynamics properly, and the performance can still be improved by adding extra nodes and connections. Although increasing the number of nodes can be an option always available at the expense of extra computational costs, the performance increase may be limited since nodes and edges are initially selected at random in the classical RC framework. Additionally, the presented pruning approach is essentially a backward truncation approach, which might not cause a significant change in the performance indices. In such cases, it is reasonable to start with a simple RC with minimum dimensionality and consider an evolutionary approach for the growth of the RC toward a more complex network architecture \cite{Yadav2024}. This evolution can be implemented by following the same greedy approach as above, but with one difference, that is, the addition of nodes and edges shall undergo a similar post-processing optimization.

\section{\label{sec:methods}Discrete-Time Extended Kalman Filter}
To provide a competitive baseline, we consider a qualitatively different filtering technique based on explicit modeling of the system dynamics. A method widely applied to nonlinear systems is EKF, a well-known tool for the denoising and reconstruction of dynamic responses from incomplete noisy measurements \cite{Anderson1979}. EKF operates on state-space models of distributed systems and requires a clear mathematical model of the system to discard noise. In Appendix A, a mathematical exposition of the EKF is provided for reference.

It should be noted that the most significant demerit of these methods lies in the fact that they require a thorough understanding of the dynamics in terms of a multi-dimensional state-space model. In contrast, the RC framework presented earlier is a model-free approach, which does not require any explicit dynamical model and provides a general mathematical framework to implicitly learn the dynamics directly from the data. In the following sections, we benchmark the results of the proposed RC framework against the EKF, showing how these methods compare in terms of validation accuracy.

\section{\label{sec:results}Results}
\subsection{Lorenz Chaotic Dynamics}
The Lorenz system is attractive to study in this work as it qualitatively exhibits different dynamics along the period-doubling route to chaos (see Appendix B for its governing equations and details). For generating training data, $\sigma = 10$, $\rho = 28$, and $\beta = 8/3$ are considered. By employing a time interval of ${\Delta}t = 0.005s$ and a recording duration of $T = 50s$, discrete-time dynamical data sets were generated considering initial conditions $x_0=y_0=z_0=1$ and polluted with specific levels of additive Gaussian White Noise (GWN). Across these data sets, the intensity of noise is considered variable as the noise standard deviation ranges from 0.25\% (52 dB) to 100\% (0 dB) of the root-mean-square (RMS) of the original signal's components. The noise-contaminated recordings of $x(t)$ and $y(t)$ are considered observables of a hypothetical measurement system and used as the input to the reservoir computer, whereas the clean version of all states is regarded as its output during training (FIG. \ref{fig:FIG. 1}). Hence, the RC must learn both denoising the inputs and reconstructing a third state. Given this specification, the generated data of $t \in [0,25 s]$ acts as the training data, and that corresponding to $t \in [25 s,50 s]$ serves as the validation set. The reservoir hyperparameters are optimized through searching within specific intervals, including the leakage rate from $\alpha \in (0.01,1)$, the spectral radius from $\gamma \in (0.01,1)$, the input scaling from $\zeta \in (0.1,2)$, and the internal connectivity of reservoir nodes from $p \in (0.1,0.9)$.

FIGS. \ref{fig:FIG. 1}\ (a) and (b) represent the inputs and outputs of the RC for reconstructing Lorenz chaotic dynamics. The results presented correspond to a reservoir computer whose hyperparameters and readout nodes are first trained, with its internal network being then truncated through the method explained above. The output of the reservoir is considerably smoother than the original noisy signals (FIG. \ref{fig:FIG. 1}). The comparison of the noisy and reproduced dynamics in phase space (FIG. \ref{fig:FIG. 1}\ (c)) shows the capability of RC in separating the dynamics of noise and responses.

\subsubsection{Optimization of Nodes and Edges}
\begin{figure}
    \centering
    \includegraphics[width=1.0\linewidth]{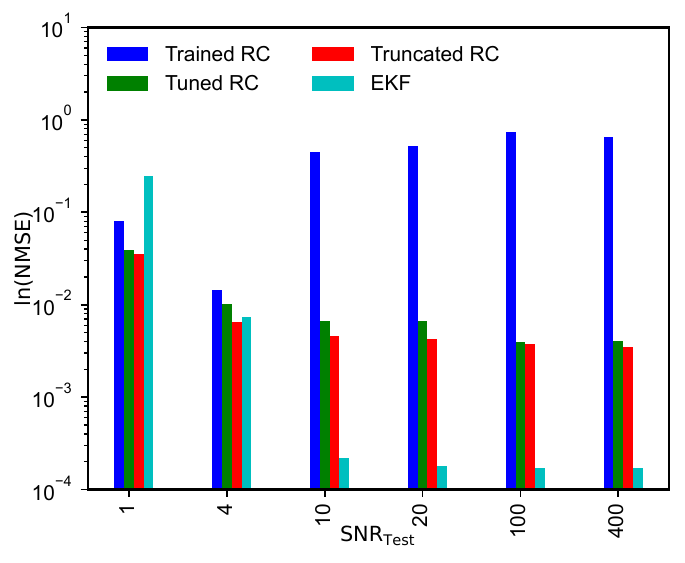}
    \caption{Performance evaluation of different training methods considering varying SNRs. Trained RC: Only the readout nodes of the RC are updated through ridge regression using 500 reservoir nodes with random edges. Tuned RC: the reservoir hyperparameters, e.g., leakage rate, spectral radius, and node connectivity ratios, were subjected to optimization while training the readout nodes as well (the number of reservoir nodes was fixed at 500). Truncated RC: The tuned RC is further optimized by reducing nodes and edges through a separate optimization.}
    \label{fig:FIG. 2}
\end{figure}

This section examines how the hyperparameter optimization and node removal approach can enhance the reservoir performance, particularly compared to a classical RC, in which the edges are random and only readout nodes are subject to calibration. The MSE is considered a measure of accuracy when the same validation set is used to compare different training schemes. However, due to the difference in the order of magnitude of the responses, a normalization over the mean-squared of the signal is included, yielding a dimensionless measure as follows:
\begin{equation}
\mathrm{NMSE} =  \frac{\|\mathbf{\hat{Y}_{\it{n}} - \mathbf{R_{\it{n}}}\mathbf{\hat{W}}_{\mathrm{out}}}\|_2^2 }{ \|\mathbf{\hat{Y}_{\it{n}}}\|_2^2}
\label{EQ6}
\end{equation}
Given the above measure, FIG. \ref{fig:FIG. 2} shows improvement in the logarithm of {\textnormal{NMSE}} for three training strategies considering different SNRs in the validation data set. Note that SNR is defined here as the RMS of the signal over the RMS of noise. The vanilla reservoir, whose readout nodes are only trained, provides the largest {\textnormal{NMSE}}, whereas further optimization of the reservoir can improve the accuracy up to several orders of magnitude in some cases. More specifically, the hyperparameter tuning presented above can considerably reduce the prediction errors, and truncating nodes and edges can further improve the accuracy. However, this superior prediction performance comes at the expense of extra computational costs.

The performance of these methods is also compared with that of the EKF. As can be seen in FIG. \ref{fig:FIG. 2}, the presented truncated RC can outperform EKF at SNRs of 1 and 4 although no explicit knowledge of the dynamics is introduced to the RC, which is contrary to the EKF that relies on the accuracy of the state-space model. Based on this result, it can be concluded that, specifically at low SNRs, the RC can offer competitive accuracy when compared with the EKF. However, it should be noted that this comparison is made while the EKF is in an advantageous position. In effect, the EKF uses physics-driven state-space models to discard noise, whereas the RC explicitly separates the desired dynamics from noise through the proposed learning protocol.

\subsubsection{Characterization of Denoising Gain}
\begin{figure}
    \centering
    \includegraphics[width=1.0\linewidth]{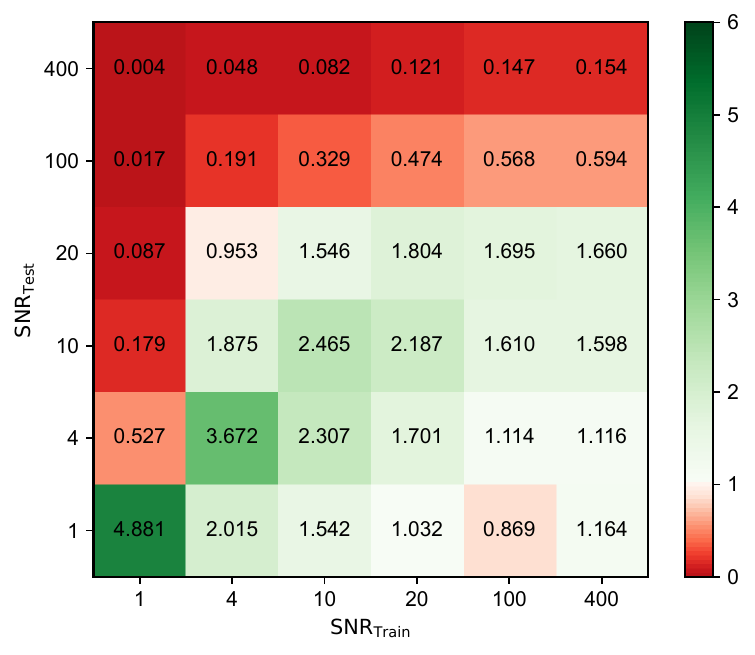}
    \caption{Denoising gain matrix obtained for training and testing data sets with different levels of noise. Note that 10 separate reservoirs are trained and tested, and the average gain values are presented to reduce the bias due to random initialization of the RC.}
    \label{fig:FIG. 3}
\end{figure}

In this section, we investigate when the reservoir is trained on a dataset with a specific SNR and how its accuracy generalizes when tested on validation datasets with smaller or larger SNRs. For this purpose, RCs and their hyperparameters are trained using datasets contaminated with varying levels of noise. Then, the calibrated RCs are used for denoising and reconstruction of held-out data sets with different levels of noise. To compare the denoising capability in these cases, a gain ratio is introduced as follows:
\begin{equation}
\mathrm{Denoising\ Gain} = \frac{\mathrm{SNR}_\mathrm{Reconstructed}}{\mathrm{SNR}_{\mathrm{Test}}}
\label{EQ7}
\end{equation}
where ${\mathrm{SNR}_\mathrm{Reconstructed}}$ represents the amount of noise present in the predictions made by the RC, and ${\mathrm{SNR}_\mathrm{Test}}$ represents the amount of noise in the test data set. When this index is greater than 1.0, it can be concluded that the RC can discard the noise, whereas values equal to or smaller than 1.0 represent cases where the RC adversely increases the amount of noise in the reconstructed data. The color-map matrix in FIG. \ref{fig:FIG. 3} visualizes the denoising gain, where the green and red colors show positive and negative denoising performance, respectively. Values on the minor diagonal of the matrix show cases where the training and testing data sets have had the same level of noise intensity. The best denoising gain is achieved when the training and testing data sets correspond to the same level of noise. Additionally, the denoising gain is better for large SNRs. For $\mathrm{SNR}_\mathrm{Train} = 100\ \textnormal{and} \ 400$ that indicates the level noise in the training set, the RC is rather unsuccessful in denoising the signals, although it still has merits in terms of reconstructing the unobserved dynamics. Additionally, the color map is completely non-symmetrical, which implies that, for a specific level of noise contained in the test dataset, those RCs trained on smaller noise perform relatively better.

\subsubsection{Generalization of Denoising Performance}
\begin{figure}
    \centering
    \includegraphics[width=1.0\linewidth]{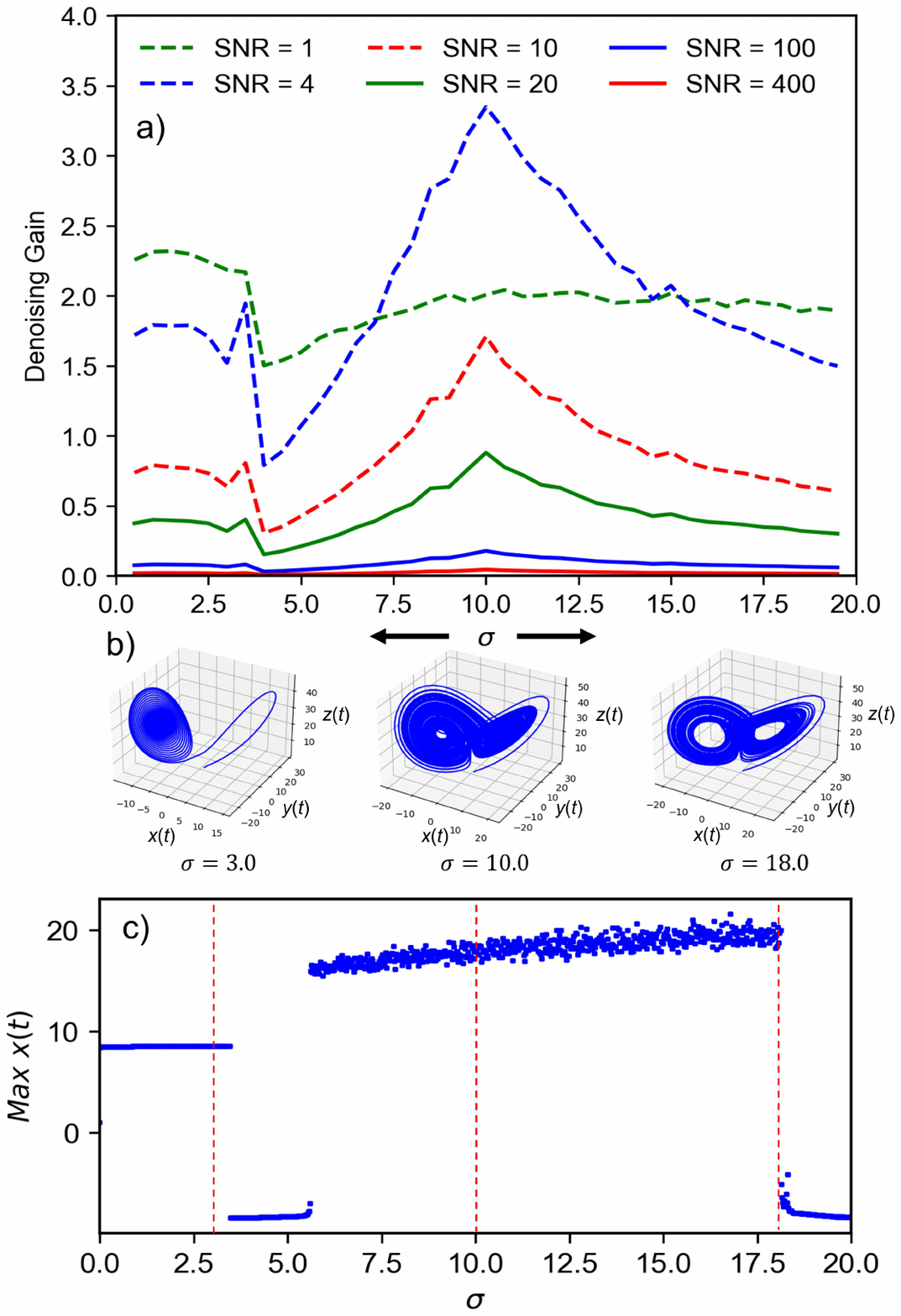}
    \caption{Generalization of denoising gain in terms of variation in Prandtl parameter. (a) Denoising gain is presented when different levels of SNR and Prandtl parameters are examined for a given RC trained on a data set with $\sigma = 10$ and SNR = 4. (b) phase space representation of Lorenz attractor for specific choices of $\sigma$. (c) Bifurcation diagram showing variation in the maximum $x(t)$. Results in sub-figure (a) are averaged over 10 randomly configured RCs to reduce the impact of randomly-initialized RCs.}
    \label{fig:FIG. 4}
\end{figure}

As demonstrated, the accuracy of the RC depends on the intensity of noise contained in the data. Similarly, it can be concluded that the accuracy might also be related to the Lorenz qualitative dynamics. To analyze such effects, the denoising gain is presented in FIG. \ref{fig:FIG. 4} for validation data sets generated with the Prandtl parameter varying within $\sigma \in [0-20]$, specifically when the RC and its hyperparameters were trained on SNR = 4 and $\sigma = 10$. As can be seen, the best gain is achieved in the vicinity of the selected Prandtl parameter ($\sigma=10$) used for training, considering the same noise level as the training data. Within a relatively wide neighborhood of $\sigma = 10$, the gain is still significantly larger than one, although the denoising gain decreases as we deviate from the center. However, there is a pattern change for values of $\sigma<4.0$, where the denoising gain increases. This observation is attributed to the gradual change in the nonlinear dynamics in the vicinity of $\sigma=4.0$, as demonstrated in the bifurcation diagram of FIG. \ref{fig:FIG. 4}(c).  This figure highlights that the denoising gain also depends on the nonlinear dynamics considered for training the RC, and it will not necessarily show monotonous behavior as we deviate from the neighborhood of the parameter used for training the RC.

\begin{figure}
    \centering
    \includegraphics[width=1.0\linewidth]{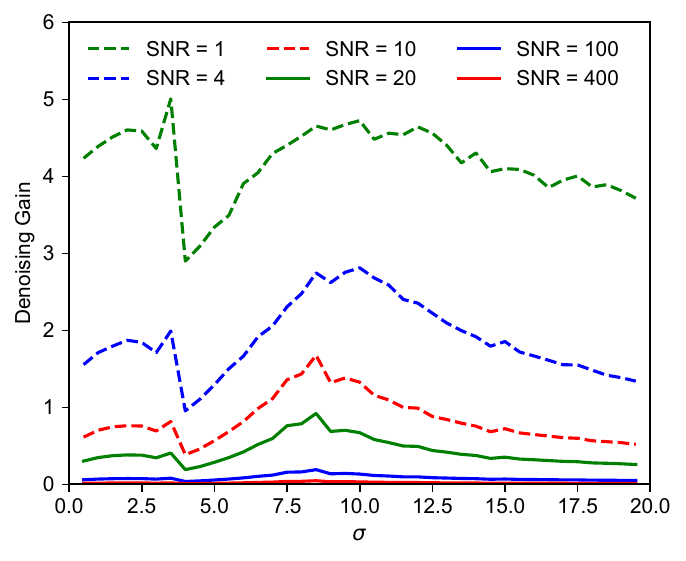}
    \caption{Improvement in denoising gain when additional data sets with $\sigma = 8$ and SNR = 4 are included in the training data. Results are averaged over 10 randomly initialized RCs.}
    \label{fig:FIG. 5}
\end{figure}

It is also worth investigating how the denoising gain varies across the domain of the Prandtl parameter, considering two additional data sets for the training of the RC and its hyperparameters. One data set is generated using $\sigma = 8$ with SNR = 1, and the other is generated using the same parameters as before, i.e., $\sigma = 10$ but using SNR = 1. Note that all other factors are considered the same as the original data set. FIG. \ref{fig:FIG. 5} presents the denoising gain for the validation data sets when the above settings are used for generating the training data. It is clear that the denoising gain considerably improves around $\sigma = 8$ and particularly for SNR = 1, exhibiting a better generalization performance across the parameter domain as well. As can be seen, by including extra training data, it is possible to enhance the denoising gain and obtain more accurate computing reservoirs. This result promotes the potential of using a reservoir network trained on a few simple chaotic orbits when applied for reconstructing unseen nonlinear dynamics and highlights how the denoising gain improves as additional data is included.

\subsection{AdEx Model}
This section examines the performance of the proposed RC framework when having stiff and non-smooth dynamics. For this purpose, the AdEx model is selected due to its bursting and adaptation behavior, which finds significant interest in studying insect locomotion and sensory threshold \cite{vanderVeen2024}. Appendix \ref{Appendix C} presents the nonlinear differential equations of this model, explaining the sensory behavior of the neurons in terms of the variation in the membrane potential $V(t)$ when driven by the current input $I(t)$. The membrane potential also depends on an independent variable, named adaptation parameter $w(t)$, simulating non-trivial firing and adaptation neuronal behavior.

\begin{figure*}
    \centering
    \includegraphics[width=1.0\linewidth]{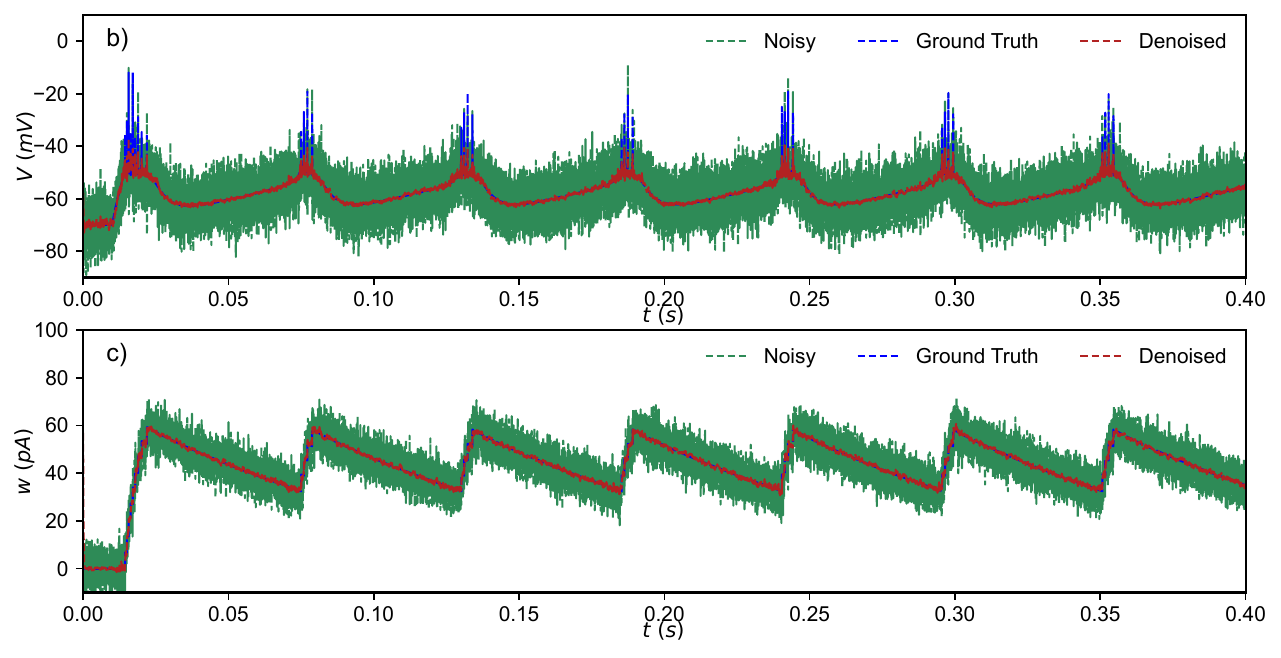}
    \caption{Reproducing noisy measurements of the AdEx model through the proposed RC framework (SNR=20 dB). The ground truth data appears in dashed blue lines, representing a neuronal behavior subjected to an initial firing pattern, followed by a subsequent adaptation. The noisy and denoised curves are shown in green and red colors. (a) Membrane potential is shown subject to fixed current input and firing patterns triggered through Dirac delta functions (b) Adaptation variable is specified in the same substance as the input current with initial firing patterns and periodic spiking behavior.}
    \label{fig:FIG. 6}
\end{figure*}

Three different noise processes are considered, varying in terms of the frequency content, comprising GWN, violet, and pink noise processes. The first 200 ms is treated as the training data, whereas the next 200 ms is considered as validation data. Reservoir nodes are not considered fixed and are released to vary between $[50-100]$ to represent cases in which the number of reservoir nodes is variable. Other hyperparameters are allowed to vary within the intervals specified earlier for the Lorenz attractor. According to Appendix \ref{Appendix C}, simulated data is generated at 100 kHz sampling rate for a period of 400 ms and polluted with 10\% RMS measurement noise (20 dB). The simulated current is considered a step function starting at 10 ms for a fixed interval of 390 ms, with a magnitude of 65 pA. Firing patterns are simulated through Dirac delta functions, implemented in the adaptation parameter equation. FIG. \ref{fig:FIG. 6} shows the ground truth membrane potential and adaptation variable along with dynamics contaminated with GWN. The results of the pruned RC are also shown in FIG. \ref{fig:FIG. 6}, presenting accurate results across both training and testing data sets when compared with the ground truth. A substantial amount of noise is discarded, and the dynamic patterns are made more evident through the presented RC framework. However, the most significant mismatch between the denoised and the ground truth membrane potential prevails at the bursting intervals, where sharp peaks are not well captured. This issue can originate from the mathematical expression of ESN, which seemingly simulates smooth dynamics better.

\begin{figure*}
    \centering
    \includegraphics[width=1.0\linewidth]{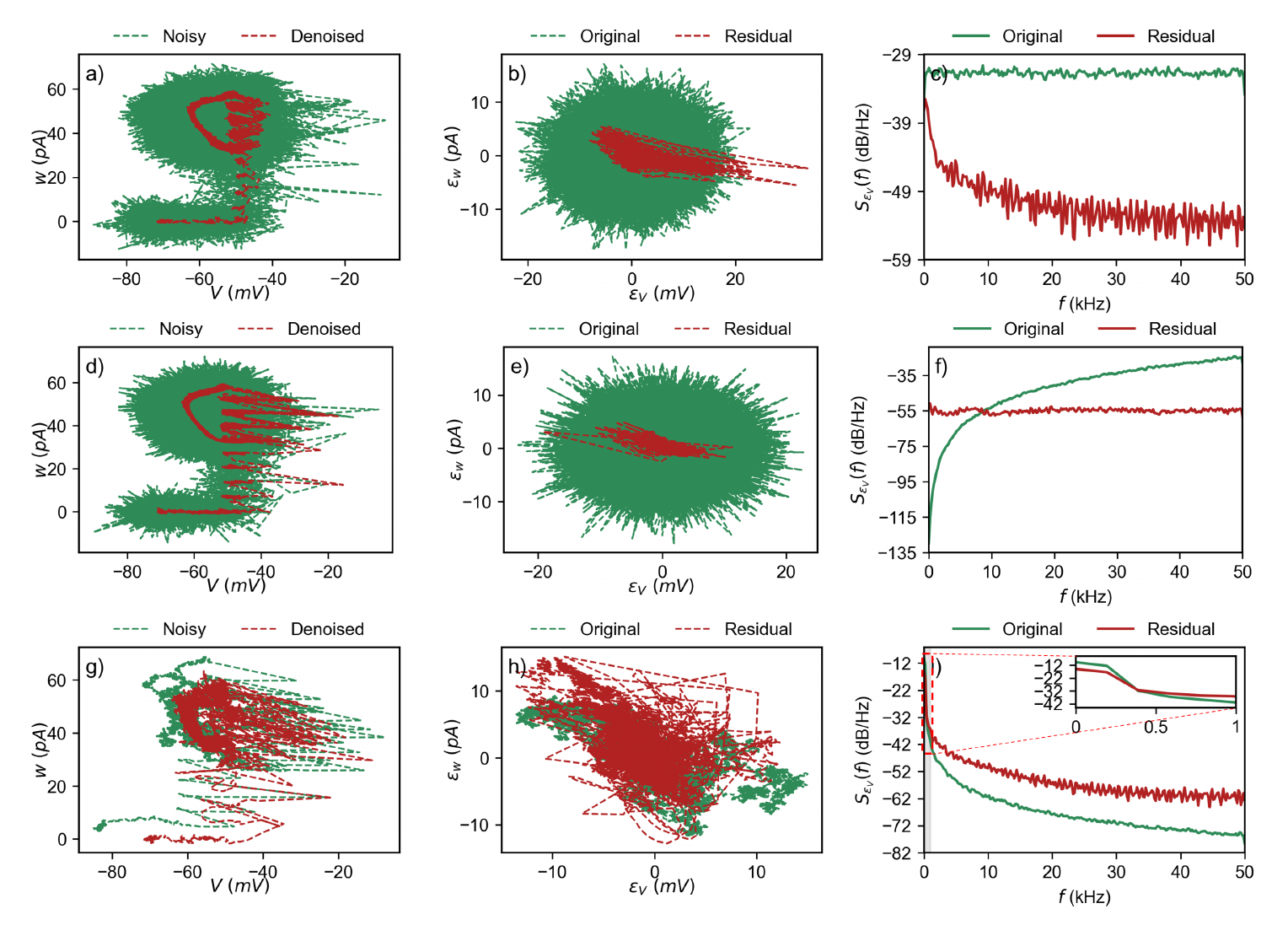}
    \caption{Denoising performance comparison between different additive noise processes (a-c) GWN $S(f)= S_0$ (d-f) Violet noise $S(f)\propto f$ (g-i) Pink noise $S(f)\propto 1/f$. The left side subplots show the denoised membrane potential and adaptation variable in the phase space compared to the noisy ones. The middle column subplots compare residual noise with the originally added noise in the phase space. The right side subplots show the averaged PSD of the residual and original noise processes.}
    \label{fig:FIG. 7}
\end{figure*}

The GWN process contains the same level of noise across the entire frequency band, but this assumption can be relaxed by considering colored noise. Thus, we investigate the denoising performance under violet and pink noise, where the high- and low-frequency components are dominant, respectively. The power spectral density (PSD) of violet noise varies linearly with the frequency, i.e., $S(f)\propto f$, whereas the PSD of pink noise is inversely proportional to the frequency, i.e., $S(f)\propto 1/f$. FIG. \ref{fig:FIG. 7} compares the denoised dynamics with the noisy ones in the phase space for the three noise processes. The denoising residual errors are also indicated in the phase space and the frequency domain for comparison purposes. These errors are denoted by $\epsilon_V(t) = V(t) -\hat{V} (t)$ and $\epsilon_w(t) = w(t) -\hat{w} (t)$, where $\hat{V} (t)$ and $\hat{w} (t)$ represent the ground truth membrane potential and adaptation parameter, respectively.

Specifically for GWN, the noise has been reduced considerably, as shown in FIG. \ref{fig:FIG. 7}(a). The subtraction of the denoised signal and the ground truth one is shown in FIG. \ref{fig:FIG. 7}(b), highlighting the capability of RC to discard GWN to a large extent. As can be seen, the errors are more pronounced along the horizontal axis. This observation can be justified based on the time-domain comparison shown in FIG. \ref{fig:FIG. 6}, where the voltage errors $\epsilon_V(t)$ are large around the sharp peaks, whereas the adaptation parameter errors $\epsilon_w(t)$ are small. Appendix \ref{Appendix D} provides further comparison between the PSD curves of the noisy and denoised signals in comparison with the ground truth PSD.

The PSD of the membrane potential error, $S_{\epsilon_{V}}(f)$, is shown in FIG. \ref{fig:FIG. 7}(c). Based on this figure, the reduction in the noise magnitude varies between 5 dB and 25 dB, depending on the frequency range. In effect, the residual noise is stronger in the low-frequency range of [0-5 kHz], but it reaches a weaker and stable regime around -50 dB/Hz at frequencies above 5 kHz, as shown in FIG. \ref{fig:FIG. 7}(c). Similar comparisons are made for violet noise in FIG. \ref{fig:FIG. 7}(d-f). The denoising performance is good. Notably, the residual noise appears similar to a GWN process with an almost fixed PSD equal to -55 dB/Hz. The original additive noise has a smaller magnitude at frequencies below 10 kHz, meaning that the RC produces extra noise in this particular frequency range. However, above 10 kHz, the denoising performance becomes significant, improving up to 30 dB/Hz.

The pink noise primarily appears as some shift in the phase space, as shown in FIG. \ref{fig:FIG. 7}(g). The RC mitigates noise at frequencies below 350 Hz, but above this range, it produces extra noise. Overall, the results display limited merits for RC in discarding pink noise.

\begin{figure}
    \centering
    \includegraphics[width=1.0\linewidth]{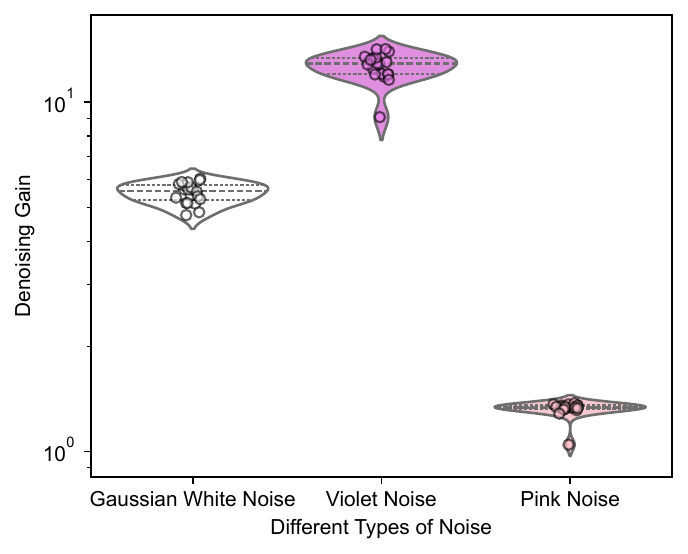}
    \caption{Distribution of denoising gain for different types of noise, showing sensitivity of the results to the initial configuration of the reservoir}
    \label{fig:FIG. 8}
\end{figure}

The results presented above correspond to one specific configuration of reservoir nodes and edges. FIG. \ref{fig:FIG. 8} shows the distribution of denoising gain (see Eq. \ref{EQ7}) when RCs are initialized with random nodal and connectivity configurations and allowed to have an optimal number of nodes between [50-100]. The results remain qualitatively consistent across different reservoir configurations. Additionally, it can be confirmed that the best denoising gain (mean: 12.68; SD: 1.13) is achieved for violet noise, wherein high-frequency components are greatly dominant. For GWN, an average denoising gain of 5.50 is acquired with a small standard deviation of 0.36. Finally, for pink noise, the average denoising gain is 1.32 with 0.07 standard deviation, which is still above 1.0, although the merits of denoising in this case are minimal.  

\section{\label{sec:discussion}Discussion}
The examples above primarily demonstrate the RC system's ability to discard additive noise. Denoising performance is evaluated using both the smooth dynamics of the Lorenz attractor and the spiking patterns of the AdEx model. While RC performs well in both scenarios, it achieves higher accuracy with smooth dynamics. This is attributed to the mathematical formulation of RCs, where the state is expressed as a linear combination of the previous state and input using a smooth activation function. 

To better handle discontinuities, alternative activation functions may be needed, ones that can approximate abrupt changes more efficiently. In this context, architectures based on spiking neuro-physical models, such as liquid state machines, could offer greater flexibility for simulating bursting nonlinear dynamics \cite{Hazan2012}. The mathematical structure of  RC appears to be problem-specific and should be tailored accordingly.

The denoising performance also varies with the SNR. For lower SNRs, the RC significantly reduces noise, yielding a denoising gain well above unity. However, at very high SNRs, the RC struggles to distinguish low-amplitude noise from dominant nonlinear dynamics, sometimes resulting in adverse denoising effects (see FIG. \ref{fig:FIG. 3}). Therefore, adequate care should be given to avoid misusing the RC.

The frequency characteristics of noise play a significant role in the denoising performance of reservoir computing. The RC demonstrates strong results when dealing with GWN and violet noise, effectively suppressing high-frequency components. However, its performance deteriorates with pink noise, where low-frequency components dominate. This limitation is expected, as low-frequency noise tends to overlap with the underlying nonlinear dynamics, making it difficult for the RC to distinguish signal from noise. This observation suggests that the denoising capabilities of RC are most effective in scenarios involving high-frequency noise at low SNR levels. Consequently, the primary hypothesis of this study requires slight refinement to reflect that RC’s denoising strengths are frequency-dependent.

These findings open up new avenues toward combining the RC with wavelet scattering \cite{Mallat2012} and less-studied nonlinear phase space-based filtering methods like the GHKSS approach \cite{Grassberger1993}. While these hybrid strategies were not exercised herein, they represent compelling directions for enhancing RC’s performance in more complex noise environments.

Another important feature of the RC is its generalization capability when the dynamical parameters change. The RC maintains its denoising gain above one across a relatively wide neighborhood of the training parameters. Including extra training data sets with varying dynamical parameters allows for acquiring higher accuracy when dealing with potential parameter changes. This adaptability is crucial for real-world applications where system dynamics may not remain constant.

While expanding the training data set improves generalization, it also increases computational costs. Specifically, when a hyperparameter optimization is required, the cost scales cubically with the dataset size due to matrix inversion in Eq. \ref{EQ4}. Therefore, careful selection of training parameters is essential to balance accuracy and computational efficiency. 

The computational cost of the ridge regression is proportional to $O(N^3)$ due to matrix multiplication for calculating the reservoir states, where $N$ is the number of reservoir nodes. The ridge regression requires a matrix multiplication and inversion of the same dimension as the reservoir matrix, having a computational cost proportional to $O(N^3)$ if the Gauss-Jordan approach is employed for the matrix inversion \cite{Golub2013}. Pruning reservoir nodes can reduce the dimensionality, and if the number of pruned nodes is $M$, the computational complexity can be characterized as $O((N-M)^3)$. This reduction can be considerably large depending on the number of pruned nodes. However, this assessment of the computational cost requires accounting for the computational efforts related to pruning the reservoir. In fact, at the training stage, the cost of pruning nodes and edges is considerable as it scales up with the number of iterations required to search for $M$ ineffective reservoir nodes.

The pruning method used to reduce the number of reservoir nodes is effective only when the node count is relatively small. The current greedy approach identifies redundant nodes based on a performance criterion like MSE, but its scalability is limited. This limitation could be improved by adopting graph-based optimization techniques, which allow for global reduction of reservoir nodes and edges without exhaustive point-by-point evaluation \cite{Scarselli2009}.

\section{\label{sec:conclusion}Conclusion}
A new RC framework was proposed for the reconstruction and denoising of nonlinear dynamics from noisy measured data. The performance of the computing reservoir was investigated in terms of the noise intensity, the sensitivity to dynamical parameters, and the noise frequency content. The optimization of the hyperparameters, the reservoir nodes, the input connectivity, the readout nodes, and the ridge regression coefficient was proven to be necessary and effective to attain a well-trained reservoir computer. Reducing the reservoir dimensionality and removing ineffective edges through a pruning technique significantly improved denoising accuracy, surpassing the performance of the classical reservoir with a fixed number of nodes and random connectivity. At high noise levels, the proposed RC outperformed an EKF implementation. This superior performance was achieved despite the fact that the EKF had an advantage as it used precise physical knowledge through the state-space model, while the RC implicitly learned the dynamics solely from the training data. Since the EKF is a near-optimal filter, having observed superior performance for the RC is of great importance from an optimality standpoint.

The denoising performance of RC was observed to be sensitive to the dynamical parameters, but the RC maintained its accuracy in the vicinity of the parameters used for generating the training data. As the validation data set deviates from the trained neighborhood of the parameters, the denoising gain drops considerably. When additional data was incorporated into the training set by using the same nonlinear dynamics but with an altered physical parameter, the denoising gain improved, showing a bi-modal pattern with two peaks positioned in the vicinity of the parameters used for training. This improvement can become the basis of training RC on dedicated training data to acquire generalization across a wider interval in the domain of physical parameters. 

The pruning technique used for reducing the dimension of the RC and eliminating redundant edges proved to be an effective approach to further reduce computational costs while enhancing accuracy. However, since the optimization was implemented based on searching for the most ineffective nodes and edges, it involves high computational costs, which scale up with the reservoir dimension. Our future works will address this problem by transitioning into a graph-based optimization algorithm, which can potentially overcome the drawbacks of searching across all nodes and edges.

\begin{acknowledgments}
This research has received funding from the Australian Research Council Linkage Project under grant number LP200301196. Additionally, the first draft of this manuscript was prepared during O. Sedehi's visit to TU Berlin, supported by the Key Technology Partnership (KTP) SEED funding scheme. This financial support is gratefully appreciated.\\

M. Stender and M. Yadav wish to acknowledge financial support from the Deutsche Forschungsgemeinschaft (DFG, German Research Foundation) under the Special Priority Program (SPP 2353: Daring More Intelligence – Design Assistants in Mechanics and Dynamics, project number 501847579).

\end{acknowledgments}

\section*{CRediT Author Statement}
O. Sedehi: Conceptualization, Methodology, Software, Validation, Formal Analysis, Investigation, Data Curation, Writing-Original Draft, Writing-Review \& Editing, Visualization; M. Yadav: Methodology, Investigation, Writing-Review \& Editing, Visualization; M. Stender: Conceptualization, Methodology, Investigation, Writing-Review \& Editing, Visualization, Supervision, Funding Acquisition; S. Oberst: Conceptualization, Methodology, Writing-Review \& Editing, Project Administration, Supervision, Funding Acquisition.

\section*{Conflict of Interest Statement}
The authors have no conflicts to disclose.

\section*{Data Availability Statement}
All codes and data that support the findings of this study are openly available in GitHub at ``https://github.com/Omid-Sedehi/ReservoirComputer.''

\begin{center}
\renewcommand\arraystretch{1.2}
\begin{tabular}{| >{\raggedright\arraybackslash}p{0.3\linewidth} | >{\raggedright\arraybackslash}p{0.65\linewidth} |}
\hline
\end{tabular}
\end{center}

\appendix
\section{\label{Appendix A}Extended Kalman filter}
It is assumed that the dynamical system under consideration can be described in discrete-time through a nonlinear state-space relationship given as
\begin{equation}
\mathbf{x}_{k} = \mathbf{g}(\mathbf{x}_{k-1}, \mathbf{u}_{k-1}) + \mathbf{v}_{k} 
\end{equation}
where ${\mathbf{x}_{k}}$ is the dynamical state at discrete time $\mathbf{\it{t}}_{k}=k\Delta t$, $\it{k}$ takes on values from $\{1,2,...,N\}$, representing individual time steps, ${\mathbf{u}_{k}}$ is the dynamical input applied at time ${\it{t_k}}$, and ${\mathbf{v}_{k}}$ is the process noise considered to be zero-mean GWN with ${\mathbf{Q}}$ covariance matrix.

Likewise, a non-linear observation model can be established by relating the state and input vectors to the observed data. This essentially leads to:
\begin{equation}
\mathbf{y}_{k} = \mathbf{h}(\mathbf{x}_{k}, \mathbf{u}_{k}) + \mathbf{w}_{k} 
\end{equation}
In this equation, ${\mathbf{y}_{k}}$ is the measured vector, ${\mathbf{h(.)}}$ is the nonlinear observation model, and ${\mathbf{v}_{k}}$ is the observation noise considered zero-mean GWN with ${\mathbf{R}}$ covariance matrix.

Through a linearization around the existing system state, it is possible to estimate the system states using the EKF. By doing so, the following sequential algorithm can be obtained \cite{Anderson1979}:
\begin{equation}
\mathbf{x}_{k|k-1} = \mathbf{g}(\mathbf{x}_{k-1|k-1}, \mathbf{u}_k)
\end{equation}
\begin{equation}
\mathbf{P}_{k|k-1} = \mathbf{F}_{k-1} \mathbf{P}_{k-1|k-1} \mathbf{F}_{k-1}^T + \mathbf{Q}
\end{equation}
\begin{equation}
\mathbf{K}_k = \mathbf{P}_{k|k-1} \mathbf{H}_k^T (\mathbf{H}_k \mathbf{P}_{k|k-1} \mathbf{H}_k^T + \mathbf{R})^{-1}
\end{equation}
\begin{equation}
\mathbf{x}_{k|k} = \mathbf{x}_{k|k-1} + \mathbf{K}_k (\mathbf{z}_k - \mathbf{h}(\mathbf{x}_{k|k-1}, \mathbf{u}_{k}))
\end{equation}
\begin{equation}
\mathbf{P}_{k|k} = \mathbf{P}_{k|k-1} - \mathbf{K}_k \mathbf{H}_k \mathbf{P}_{k|k-1}
\end{equation}

Here, $\mathbf{x}_{k|k-1}$ is the predicted state, $\mathbf{x}_{k|k}$ is the estimated state, $\mathbf{P}_{k|k-1}$ is the prediction error covariance matrix, $\mathbf{P}_{k|k}$ is the updated error covariance matrix, $\mathbf{K}_k$ is the Kalman gain matrix, $\mathbf{F}_{k-1}$ is the derivative of $\mathbf{g(.)}$  with respect to $\mathbf{x}$ evaluated at $\mathbf{x}_{k-1|k-1}$, $\mathbf{H}_k$ is the derivative of the observation model \( \mathbf{h(.)} \) with respect to \( \mathbf{x_k} \) evaluated at \( \mathbf{x}_{k|k-1} \).

Given the noise covariance matrices, the above sequential formulations provide an efficient estimate of the system states. 

\section{\label{Appendix B}Nonlinear equations of Lorenz system}

The Lorenz system consists of three major system state variables, which govern the atmospheric convection through the following set of equations \cite{Lorenz1963}:
\begin{align}
\frac{dx}{dt} &= \sigma (y - x) \\
\frac{dy}{dt} &= x (\rho - z) - y \\
\frac{dz}{dt} &= xy - \beta z
\end{align}
where $\sigma$ is the Prandtl number, $\rho$ is Rayleigh number, and $\beta$ is dissipation rate. Note that selecting different parameters can entirely change the dynamic regimes. The dynamics of the Lorenz system are nonlinear, chaotic, and smooth deterministic, which depend heavily on the initial conditions.

\section{\label{Appendix C}Mathematical formulation of AdEx model}
The nonlinear differential equations describing the adaptive exponential integrate-and-fire model are provided in the following \cite{gerstner2014neuronal}:
\begin{align}
\tau_m C\frac{dV}{dt} &= -(V-V_r) + \Delta_T \exp(\frac{V-V_T}{\Delta_T}) - Rw + RI\\
\tau_w \frac{dw}{dt} &= a(V-V_r) - w + b\tau_w\sum_{t_f}\delta(t-t_f)
\end{align}
where $V$ is voltage, $I$ is input current, $w$ is adaptation parameter, $V_r$ is leak reversal potential, $t_f$ is firing times at which the voltage exceeds $V_T$ threshold, $\delta(.)$ is Dirac delta function. Other parameters are constant coefficients, including $\tau_m = 5ms$ is membrane time scale, $\tau_w = 100ms$ is adaptation time constant, $R=500M\Omega$ is membrane resistance, $V_r=-55mV$ is rest potential, $V_T=-51mV$ is voltage threshold, $\Delta_T=-2mV$ is sharpness threshold, $a=-0.5ns$ is adaptation voltage coupling, and $b=7pA$ is spike-triggered adaptation increment. At spike time $t = t_f$, there will be a sharp rise in the adaptation parameter equal to $b$, as described via the Dirac delta function. The dynamics of the AdEx model are nonlinear and can be chaotic, showing non-smooth deterministic behavior, owing to the threshold-reset mechanism.

\section{\label{Appendix D}Comparison of PSD curves in AdEx example}
\begin{figure}
    \centering
    \includegraphics[width=1.0\linewidth]{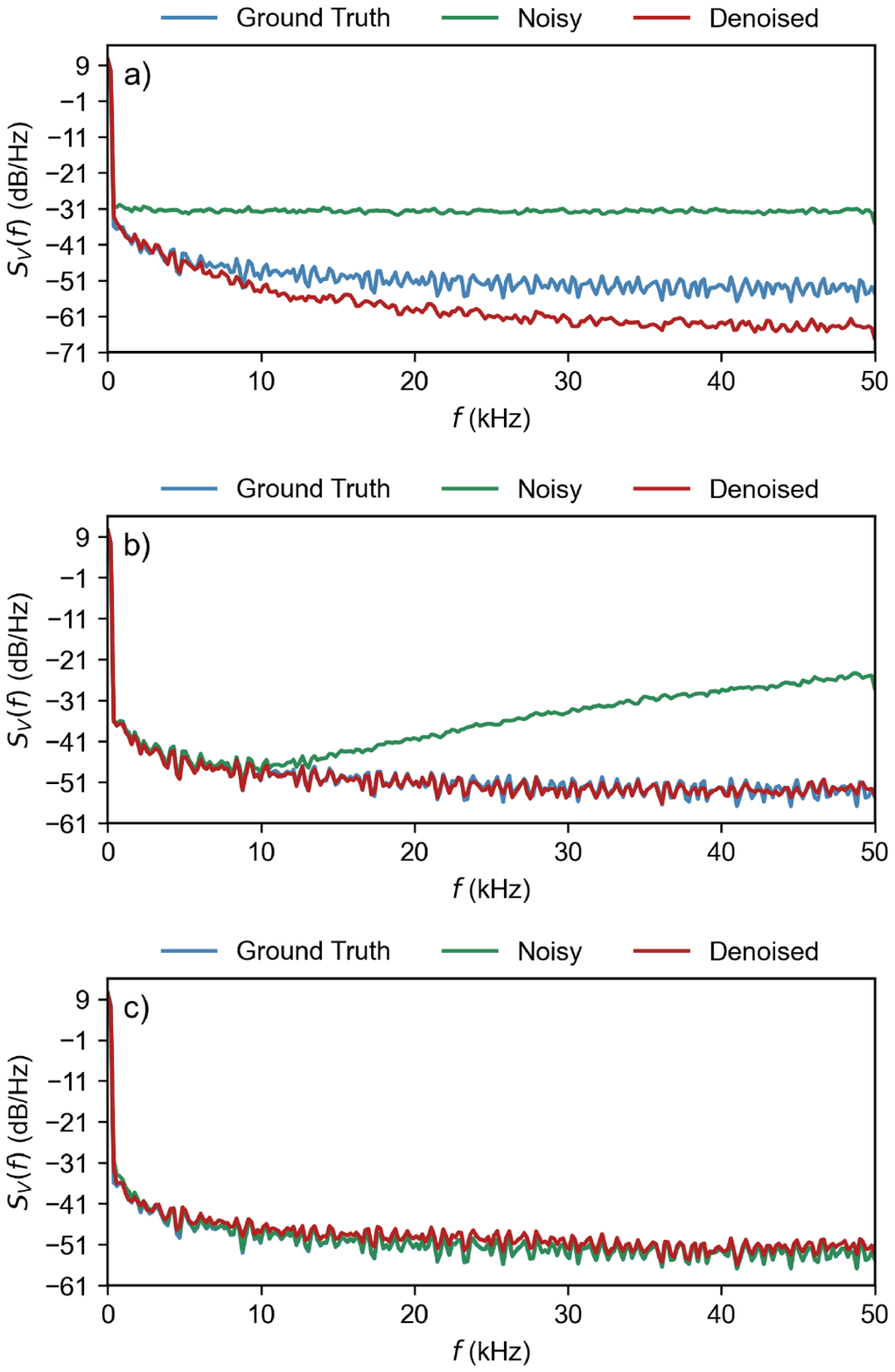}
    \caption{PSD curves of the membrane potential when considering additive (a) GWN (b) Violet noise (c) Pink noise. ``Ground Truth'' corresponds to the noise-free original signal; ``Noisy'' represents the noisy signal fed into the RC framework; and ``Denoised'' represents the reconstructed signal.}
    \label{fig:FIG. 9}
\end{figure}

The results presented in FIG. \ref{fig:FIG. 7} compares the PSD of the membrane potential prediction errors, i.e., $S_{\epsilon_V} (f)$. FIG. \ref{fig:FIG. 9} supplements those results, showing the PSD of the ground truth membrane potential, $S_V (f)$, compared to those of the noisy and denoised signals. Note that the difference of the noisy and denoised signals from the ground truth signal is indicated in the PSD curves of FIG. \ref{fig:FIG. 7} and labeled as ``original'' and ``residual.'' These results further confirm the above conclusions, showing the performance of RC in handling different types of noise and emphasizing that the RC is more successful in discarding high-frequency noise.

\bibliography{aipsamp1}

\end{document}